\documentclass[conference]{IEEEtran}
\IEEEoverridecommandlockouts
\usepackage{cite}
\usepackage{amsmath,amssymb,amsfonts}
\usepackage{graphicx}
\usepackage{textcomp}
\usepackage{xcolor}

\usepackage{hyperref}
\usepackage{siunitx}
\usepackage{booktabs}
\usepackage{amsmath,mathrsfs}
\usepackage{mathtools}
\usepackage{amssymb}
\usepackage{graphicx}
\usepackage[export]{adjustbox}
\usepackage{float}
\usepackage{algorithm}
\usepackage[noend]{algpseudocode}
\usepackage{overpic}

\graphicspath{ . }

\newcommand{\ve}[1]{\ensuremath{\mbox{\boldmath$#1$}}}

\usepackage{listings}

\def\BibTeX{{\rm B\kern-.05em{\sc i\kern-.025em b}\kern-.08em
    T\kern-.1667em\lower.7ex\hbox{E}\kern-.125emX}}
    \newcommand{\citep}{\cite}
\begin{document}

\title{Improving traffic sign recognition by active search
\\
}

\author{
\IEEEauthorblockN{
S. Jaghouar\IEEEauthorrefmark{1},
H. Gustafsson\IEEEauthorrefmark{2}, 
B. Mehlig,\IEEEauthorrefmark{3}, 
E. Werner\IEEEauthorrefmark{4}, 
N.Gustafsson\IEEEauthorrefmark{5}
}
\IEEEauthorblockA{
\IEEEauthorrefmark{1} \textit{University Of Technology of Compiegne}, Compiegne, France, sami.jaghouar@gmail.com\\
\IEEEauthorrefmark{2}\textit{Chalmers University of Technology},  Gothenburg, Sweden, {ghannes@student.chalmers.se}\\
\IEEEauthorrefmark{3}\textit{University of Gothenburg}, Gothenburg, Sweden, bernhard.mehlig@physics.gu.se\\
\IEEEauthorrefmark{4}\textit{Zenseact}, Gothenburg, Sweden, erik.werner@zenseact.com\\
\IEEEauthorrefmark{5}\textit{Zenseact}, Gothenburg, Sweden, niklas.gustafsson@zenseact.com
}}

\maketitle

\begin{abstract}
We describe an iterative active-learning algorithm to recognise rare traffic signs. A standard ResNet is trained on a training set containing only a single sample of the rare class. We demonstrate that by sorting the samples of a large, unlabeled set by the estimated probability of belonging to the rare class, we can efficiently identify samples from the rare class. This works despite the fact that this estimated probability is usually quite low. A reliable active-learning loop is obtained by labeling these candidate samples, including them in the training set, and iterating the procedure. Further, we show that we get similar results starting from a single synthetic sample. Our results are important as they indicate a straightforward way of improving traffic-sign recognition for automated driving systems. In addition, they show that we can make use of the information hidden in low confidence outputs, which is usually ignored.
\end{abstract}


\section{Introduction}
\label{introduction}

\begin{figure}[t]
        \includegraphics[width=0.9\columnwidth]{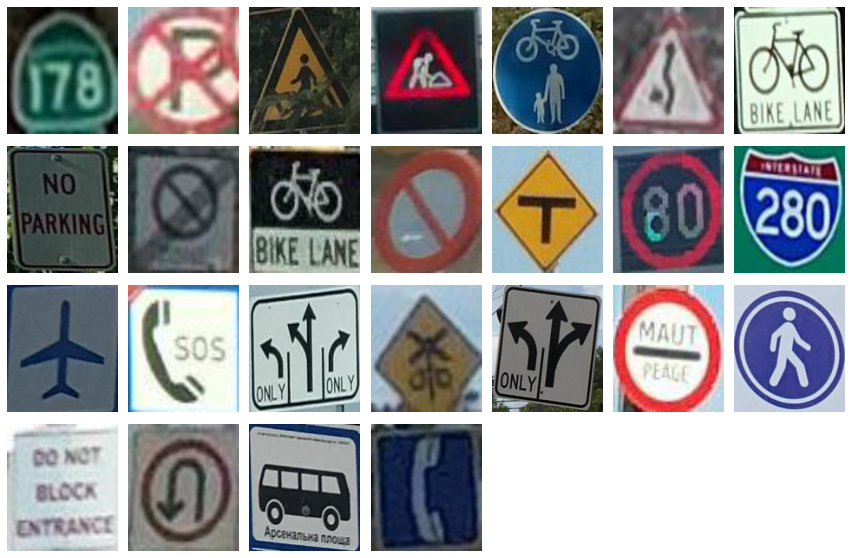}
        \caption{Samples of the 25 rarest traffic signs in the Mapillary data set \citep{ertler_mapillary_2020}}
        \label{fig:25_rarest_class}
        \centering
\end{figure}

Deep neural networks are now the standard choice for perception systems in self-driving cars \citep{yurtsever_survey_2020}. However, they can perform poorly on {\em rare} classes, for which there are few samples in the training set. This is problematic for traffic-sign recognition, since some signs occur much less often than others. As an example, Fig.~\ref{fig:25_rarest_class} shows samples from the 25 rarest classes in the  Mapillary traffic-sign data set \citep{ertler_mapillary_2020}.

One way to improve the performance is to find more samples of these rare signs in the raw data, label them, and add them to the training set. However, finding the rare traffic signs in the raw, unlabeled dataset is a challenge in itself, a task that has been likened to finding a needle in a haystack~\citep{coleman_similarity_2020,yue2021interventional}.

In this paper, we describe an active-learning algorithm for finding rare traffic signs in a large unlabeled data set.
The algorithm is based on a standard ResNet \citep{he_deep_2015}, trained on a heavily imbalanced training set, which contains one sample of each rare class from Fig.~\ref{fig:25_rarest_class}.
The algorithm finds more samples of each rare sign in a large set of unlabeled data, by simply selecting the samples with the highest estimated probability of belonging to the rare class.

\begin{figure}
\mbox{}\hspace*{10mm}
\begin{overpic}[width=0.125\columnwidth]{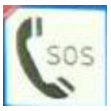}
\put(-40,73){({\bf a})}
\end{overpic}\\[0.2cm]
\mbox{}\hspace*{9.5mm}
        \begin{overpic}[width=0.8\columnwidth]{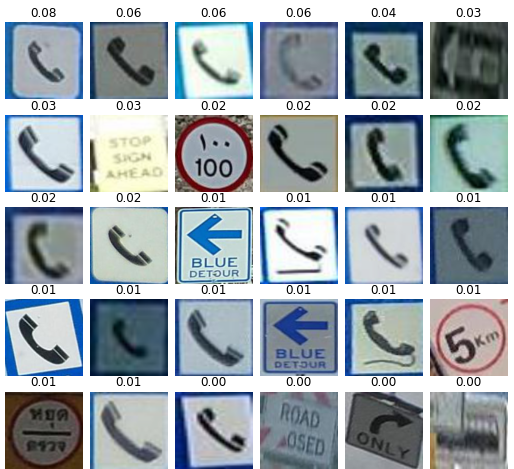}
\put(-6,83){({\bf b})}
\end{overpic}
        \caption{An illustration of an essential step in the algorithm to find rare traffic signs, for the sign shown in panel ({\bf a}). Panel ({\bf b}) shows the 30 signs in the unlabeled data set with the highest probability of belonging to the same class as this sign, as estimated by the neural network. This probability is indicated above each sample. }
\label{fig:qualitative_searching}
\end{figure}

This works quite well, despite the fact that this estimated probability is often low. We show that the low probabilities estimated by the network nevertheless contain important information -- sorting the unlabeled set by this probability reveals that a large fraction of the highest scoring samples belongs to the rare class (Fig.~\ref{fig:qualitative_searching}). By labeling these samples, and adding them to the training set, we can rapidly improve the performance of our model on the rare classes. This algorithm works well even when the single sample from each rare class is a synthetic image.

The motivation for this research was to improve the recognition of rare traffic signs for automated driving. Apart from the practical significance of our results, an interesting finding of our study is that it is possible to make use of the information present in very low estimated probabilities.

\section{Related work}
\label{related-work}

\textbf{Active learning}
describes a collection of iterative methods to expand the training set for neural networks. Since manual labeling is expensive, the goal of active learning
is to use the output of the network to find the most useful samples to add to the training set \citep{ren_survey_2020}.
A common approach is the pool-based active learning loop \citep{Budd_2021}. The network is trained on a labeled training set, and used to select samples from an unlabeled dataset, or  {\em pool}, commonly via uncertainty sampling using e.g. entropy \citep{ren_survey_2020}. The samples are labeled and added to the training set. This loop can be repeated a number of times.

The process of searching the unlabeled dataset for rare samples,
in order to add them to the training set to improve performance on those classes, is called {\em active search} \citep{jiang_efficient_2017,jiang_cost_2019,coleman_similarity_2021}.
While active learning has been used to tackle the traffic sign recognition problem \citep{inproceedings}, to the best of our knowledge there is no published research on active search for rare traffic sign recognition.

\textbf{Few-shot learning} algorithms can learn to recognize an image from only few examples \citep{wang_generalizing_2020}. The algorithm can grasp the essential features of a sample and generalise them, enabling the algorithm to correctly classify e.g. traffic signs it has only seen a few times before, perhaps even only once. These algorithms tend to follow the meta-learning paradigm \citep{ren_meta-learning_2018-1,finn_model-agnostic_2017}, meaning they {\em learn to learn} by training on small classification problems called episodes. The algorithms commonly rely on a feature extractor that represents the essential traffic-sign features in a high-dimensional feature space. Distance based techniques  \citep{snell_prototypical_2017,sung_learning_2018,vinyals_matching_2017} or smart optimisers \citep{andrychowicz_learning_2016} are then used to interpret clusters in feature space in order to correctly classify rare traffic-sings.

\textbf{Class imbalance} is a well-known problem in machine learning. A classification model may achieve excellent overall performance, e.g. accuracy, yet fail on classes which are underrepresented in the training set. A common explanation for this is that the few rare samples will make a small contribution to the total training loss \citep{johnson_survey_2019}. One could use oversampling \citep{buda_systematic_2018} to artificially increase this contribution. Instead, we show that one can find meaningful information in the low softmax outputs for the rare classes.

\section{Methods}
\label{methods}

\subsection{Datasets}
\label{data_sets}
\begin{figure}[t]
\centering
 \begin{overpic}[width=0.6\columnwidth]{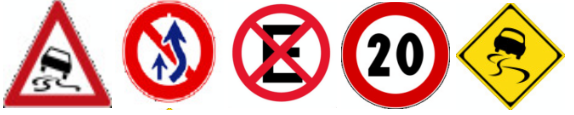}
\put(-7,15.5){({\bf a})}
\end{overpic}\\[4mm]
\mbox{}\hspace*{1.75mm}\begin{overpic}[width=0.6\columnwidth]{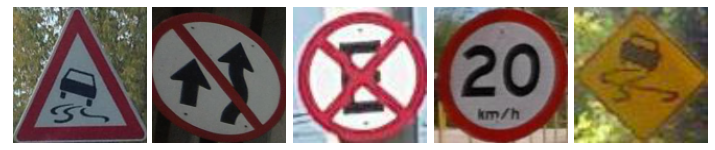}
\put(-8,15.5){({\bf b})}
\end{overpic}
        \caption{({\bf a}) Synthetic samples of rare traffic signs from wikipedia \citep{wiki_traffic_sign}. ({\bf b})~Corresponding signs from the Mapillary data set.}
        \label{fig:6_rare_class}
        \centering
\end{figure}

\begin{algorithm}[t]
  \caption{Python-style pseudocode for our active learning algorithm}
  \label{algo:active_loop}
    \definecolor{codeblue}{rgb}{0.25,0.5,0.5}
    \definecolor{codekw}{rgb}{0.85, 0.18, 0.50}
    \newcommand{\algofontsize}{8pt}
    \lstset{
      backgroundcolor=\color{white},
      basicstyle=\fontsize{\algofontsize}{\algofontsize}\ttfamily\selectfont,
      columns=fullflexible,
      breaklines=false,
      captionpos=b,
      commentstyle=\fontsize{\algofontsize}{\algofontsize}\color{codeblue},
      keywordstyle=\fontsize{\algofontsize}{\algofontsize}\color{purple},
      numberstyle =\fontsize{\algofontsize}{\algofontsize}\color{red},
    }

\begin{lstlisting}[language=python]
# training_set: initial training set
# unlabeled_set: initial unlabeled set
# validation_set: initial validation set
# rare_classes: list of rare class indices
# model: neural network
# N: #frames to label per class in each iteration
# T: #iteration steps

model.fit(training_set)

F1 = []
F1.append(f1_metrics(model, validation_set))

for t in range(T): # active-learning loop
    for i in rare_classes:
        scores = [model(x)[i] for x in unlabeled_set]
        indices = scores.argsort()[:N]
        selected_frames = unlabeled_set[indices]
        labeled_frames = label(selected_frames)
        training_set.extend(labeled_frames)
        unlabeled_set.remove(selected_frames)

    model.fit(training_set)
    F1.append(f1_metrics(model, validation_set))
\end{lstlisting}
\end{algorithm}

The Mapillary dataset \citep{ertler_mapillary_2020} was used in all of our experiments. The dataset consists of images of scenes with traffic signs in them. The bounding box for each sign was used to extract all image patches with a traffic sign. All patches smaller than 30x30 pixels were discarded, resulting in a labeled dataset of 59150 traffic signs, belonging to 313 different classes. Lastly, all patches were reshaped to 128x128 pixels.

The data was split into three sets. A training and validation set,
and a third set to use as an unlabeled dataset. The training set contained 18306 samples, and the validation set contained 2560 samples. The remaining 38284 traffic signs were used as a pool of unlabeled traffic signs.

The 25 rarest traffic-sign classes shown in Fig.~\ref{fig:25_rarest_class} were used to test the algorithm. For each rare class a single, randomly selected, sample was added to the training set, while two to three randomly selected ones were added to the validation set. The remaining ones, between four and fifty samples per class, were added to the unlabeled set.

Experiments were also performed using synthetic data. In these experiments, 
a single synthetic sample [Fig.~\ref{fig:6_rare_class}({\bf a})] corresponding
to one of the five rare classes shown in Fig.~\ref{fig:6_rare_class}({\bf b})
was added to the training set instead of a randomly selected sample from the dataset. We limited the number of samples in the unlabeled dataset to fifty for each rare class.

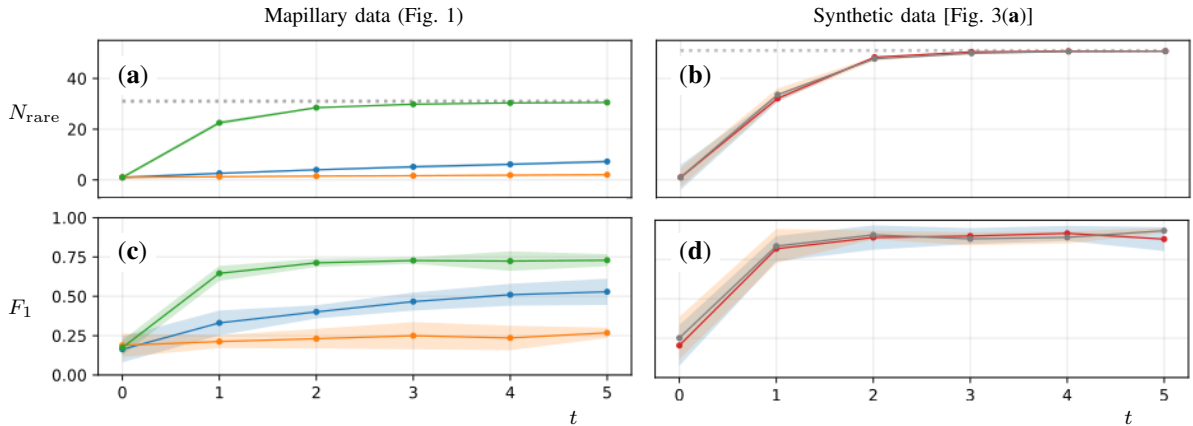
\begin{figure*}[t]
\hspace*{15mm}
\begin{overpic}[scale=0.6,angle =0]{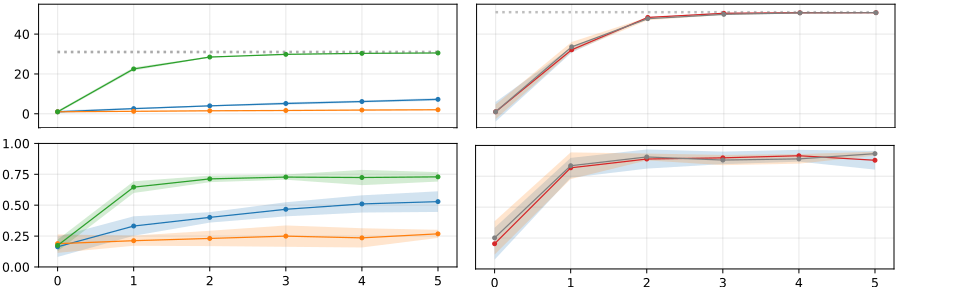}
\put(5,27){\colorbox{white}{({\bf a})}}
\put(51,27){\colorbox{white}{({\bf b})}}
\put(5,12.5){\colorbox{white}{({\bf c})}}
\put(51,12.5){\colorbox{white}{({\bf d})}}
\put(-4,24){\colorbox{white}{\footnotesize $N_{\rm rare}$}}
\put(-4,8){\colorbox{white}{\footnotesize $F_1$}}
\put(42,-1){\colorbox{white}{\footnotesize $t$}}
\put(87.5,-1){\colorbox{white}{\footnotesize $t$}}
\put(17.,32){\colorbox{white}{\footnotesize Mapillary data (Fig.~\ref{fig:25_rarest_class})}}
\put(62,32){\colorbox{white}{\footnotesize Synthetic data [Fig.~\ref{fig:6_rare_class}({\bf a})]}}
\end{overpic}
        \caption{Performance of the active-learning algorithm.
({\bf a}) The number $N_{\rm rare}$ of rare samples in the training set, averaged over the rare classes in Fig.~\ref{fig:25_rarest_class}, as a function of the iteration number $t$ of the active loop. Shown are results of our active-learning algorithm (green); entropy method (Section~\ref{benchmark}), blue; random selection (Section~\ref{benchmark}), orange. The dotted gray line shows the maximal value, i.e. when all the rare samples have been moved to the training set.
({\bf c}) Same, but for the $F_1$ score of the rare classes. In panels ({\bf b}) and ({\bf d}), we show the corresponding results, but using synthetic samples for the five rare classes in Fig.~\ref{fig:6_rare_class} (black). For comparison, we also show the result when using Mapillary samples in Fig. \ref{fig:6_rare_class}  (red), instead of synthetic ones.  In each panel, we report the results with 95\% confidence intervals obtained from five runs.}
	\label{fig:25_rare_al_50_1} \centering
        \centering
\end{figure*}

\subsection{Neural-network model}
\label{nn}
In all experiments, the torchvision \citep{tv} implementation of ResNet18 \citep{he_deep_2015} was used. The network was pretrained on ImageNet \citep{imagenet}, except for the last classification layer which was randomly initialized.
Random cropping was applied as data augmentation. The network was trained using the ADAM optimizer \citep{kingma_adam_2017} with a learning rate of $10^{-3}$ and a batch size of 256 for 20 epochs.

\subsection{Active-learning algorithm}
\label{ala}
One iteration of the active-learning algorithm consists of three steps.
First, the network is trained on the training set.
Second, the network is used to predict the class of each unlabeled sample. Lastly, for each rare class, the samples are sorted in descending order using the estimated probability of belonging to the rare class as shown in Fig.~\ref{fig:qualitative_searching}, and the $N$ first samples are labeled and moved to the training dataset.

The algorithm is summarised in Algorithm \ref{algo:active_loop}, and our implementation is available on \href{https://github.com/samsja/finding-a-needle-code}{github}. Two experiments were conducted, one using the rare classes from Fig.~\ref{fig:25_rarest_class} and one on synthetic data using the rare classes shown in Fig.~\ref{fig:6_rare_class}. In these experiments $N$ was set to 50, resulting in $50\times K$ samples being moved from the unlabeled dataset to the training dataset in each iteration, where $K$ is the number of rare classes. This means $50 \times 25 = 1250$ samples were moved in the experiment using only real data, while $50 \times 5 = 250$ samples were moved in the experiments on synthetic data.

\subsection{Evaluation of performance}
\label{s-score}

The $F_1$ score on the rare classes was used to evaluate the performance of the algorithm described in Section \ref{ala}. In addition, the number of training samples per class averaged over all of the rare classes was computed. This is denoted by $N_{\rm rare}$. At the first iteration we have $N_{\rm rare} = 1$, since there is one sample for each rare class in the training set. When rare samples are found and added to the training set, $N_{\rm rare}$ increases.

\subsection{Benchmarks}
\label{benchmark}
The results of our active-learning algorithm are compared against random selection and a commonly used active-learning method - selecting the samples with the highest entropy \citep{ren_survey_2020}.
In the Appendix, we also show the result of applying two few-shot learning techniques \citep{snell_prototypical_2017,sung_learning_2018}  to our problem.

\section{Results}
\label{results}

\label{active-loop}

Fig.~\ref{fig:25_rare_al_50_1} illustrates how the algorithm iteratively finds more and more samples of the rare class. The figure shows $N_{\rm rare}$, the number of training samples per class averaged over the rare classes from Fig.~\ref{fig:25_rarest_class}, as well as the $F_1$ score. Both quantities are plotted versus the iteration number $t$ of the active loop (see Section~\ref{benchmark}). Also shown are the results of selecting the samples with highest entropy and random selection (Section \ref{benchmark}). The new algorithm significantly outperforms the other two methods.

The value of $N_{\rm rare}$ tells us how many new rare samples were added in each iteration of the active loop.
Fig.~\ref{fig:25_rare_al_50_1}({\bf a}) shows that in the first iteration, our method finds on average 22 new training samples per rare class in the unlabeled dataset. The two other two algorithms, by contrast, find fewer than five of them, and after five iterations, $N_{\rm rare}$ remains below ten for both of these algorithms.

Now consider the $F_1$ score [Fig.~\ref{fig:25_rare_al_50_1}({\bf c})]. For our algorithm, the $F_1$ score on the rare classes increases from 0.17 for the initial training set, to 0.72 after the third active learning iteration. For the entropy and the random-selection methods, the $F_1$ score for the rare classes increases more slowly [Fig.~\ref{fig:25_rare_al_50_1}({\bf c})].

In the first two iterations of our algorithm, our method finds more and more rare samples to add to the training set. Naturally, this helps increase the recognition performance, as measured by the $F_1$ score on the rare classes. However, the $F_1$ score starts to plateau after the first two iterations. The reason is that our algorithm  manages to find almost all of the  rare samples in the unlabeled dataset in the first two iterations.  Therefore, in the later iterations it is impossible to add more than
a small number of new rare samples, and as a result there is only a small improvement for the last three iterations. 
The confidence intervals in Fig.~\ref{fig:25_rare_al_50_1}({\bf a},{\bf c}) are quite small. This indicates that the method is quite robust.

Fig.~\ref{fig:25_rare_al_50_1}({\bf b},{\bf d}) compares the performance for two different initializations: using a rare sample from the Mapillary data set in the training set ({red}), and using a synthetic sample ({black}) [Fig.~\ref{fig:6_rare_class}({\bf b})] .
The figure shows that it does not matter much whether the algorithm first sees an actual Mapillary sample or a synthetic one, it works as efficiently.

 Note that the final $F_1$ score is higher in Fig.~\ref{fig:25_rare_al_50_1}({\bf d}) than in Fig.~\ref{fig:25_rare_al_50_1}({\bf c}). This indicates that the 5 classes for which there are synthetic samples [Fig.~\ref{fig:6_rare_class}({\bf a})] are on average easier to recognize than the 25 classes used in the non-synthetic experiment (Fig.~\ref{fig:25_rarest_class}). Moreover, these 5 classes have more samples in the unlabeled dataset than the 25 rarest classes, and thus at the end of the fifth iteration there are more training samples for these classes [Fig.~\ref{fig:25_rare_al_50_1}({\bf b})].

In the Appendix, we show the performance of our active-search algorithm, when the standard supervised training is replaced by two few-shot learning methods. Contrary to what one might expect, there is no performance gain when incorporating these techniques. This is in line with recent work indicating that standard neural networks may do just as well at few-shot learning, and that what really matters is the quality of the feature extraction \citep{chowdhury_few-shot_2021,chen_closer_2020,kolesnikov_big_2020,tian_rethinking_2020,zhai_scaling_2021,raghu_rapid_2019}.

\section{Discussion}
\label{discussion}

Fig.~\ref{fig:qualitative_searching} shows something unintuitive.
The classification network does a very good job at finding samples from the rare class -- e.g. 16 out of the top 20 belong to this class.
However, the probability assigned to this class is always very low, below 0.1 for all samples.
In this Section we explain why there is relevant information in the small softmax outputs, and how it can be used.

We can analyse the classification of a given input $\ve x$ in two steps.
First, the ResNet model maps the input $\ve x$ to a 512-dimensional feature vector $\ve z$. Then, the final layer of ResNet maps the feature vector to a probability estimate by passing it through a softmax function,
\begin{align}
	P_i = \frac{\exp(\ve W_i \cdot \ve z + b_i)}{\sum_j \exp(\ve W_j  \cdot \ve z + b_j)}\,.
	\label{eq:linear_output_model}
\end{align}
Here $P_i$ is the estimated probability that the input $\ve x$ belongs to class $i$. $\ve W_i $ is the weight vector of the output unit corresponding to the rare class, and $b_i$ is its bias.

Consider the output for a rare class.
If the network is trained with a normal cross-entropy loss, as in our case, the rare class can only make a limited contribution to the loss function.
As a result, the rare class has a quite limited influence on the feature vector $\ve z$.
Further, the network quickly learns to output a small probability for the rare class, for all samples with a different label.
E.g. by assigning a large negative value to $b_i$. However, to avoid a very high loss on the few samples with the rare label, the network learns to minimise this loss by updating $\ve W_i $ to point along the direction in feature space that gives the highest probability for class~$i$.
In other words, even if the probability $P(y_i)$ is small for all samples in the training set, it will still be largest for those samples that have the label $i$.

If the features of the unseen samples of the rare class are also in the direction of $\ve W_i $, the new active-learning algorithm has a large chance of finding these samples in the unlabeled data set.
\begin{figure}[t]
	\begin{overpic}[width=0.45\textwidth]{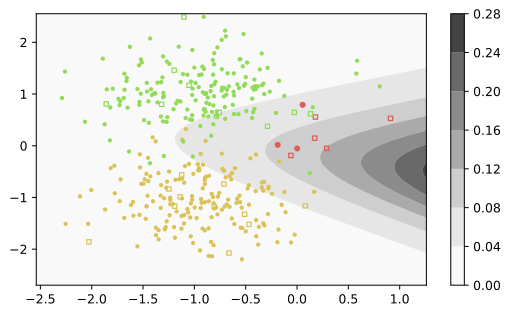}
        \put(52,0){$z_1$}
        \put(0,48){$z_2$}
        \end{overpic}
	\caption{A toy model of a feature space with one rare class (red points), and two common ones (yellow and green points). We trained a one-layer perceptron with cross-entropy loss, and show a contour plot of the estimated probability that a point belongs to the rare class. The network is trained on all inputs represented by filled symbols. The other points (empty symbols) represent the unlabeled dataset. }
	\label{fig:toy_model}
	\centering
\end{figure}
To illustrate this point, we consider a toy model for classification [Fig.~\ref{fig:toy_model}], with one rare class and two common ones in two-dimensional input space.
The samples of each class are generated by Gaussian distributions $\mathcal{N}(\mu, \sigma^2)$, where $\mu = ({-}1, \pm 1)$ for the common classes, $\mu = (0, 0)$ for the rare class,
and $\sigma=0.5$.
This two-dimensional space is a highly idealised model of the feature space of a deep neural network.

On this toy dataset we train a linear classifier with a cross-entropy loss.
The color map of Fig.~\ref{fig:toy_model} shows the estimated probability for the rare class, Eq.~\ref{eq:linear_output_model}, for each point in input space.
We want to highlight three aspects of this output:
First, the estimated probability is small ($\le 0.2$) for all input samples.
Second, it is largest for inputs in the right-hand part of the plane, corresponding to the direction of the rare cluster.
Third, despite the large class imbalance, if we were to sort all unlabeled samples according to this probability, a large fraction of the highest scoring ones would belong to the rare class.

In the Appendix, we show that the feature space of our ResNet model is similar to this toy model. In particular, we show that the feature space is expressive enough that all samples of a rare class are in a consistent direction in feature space.

\section{Conclusions}
\label{conclusion}

In this work, we considered the problem of how to learn to classify rare traffic signs. Our main result is that we obtain very good results using a simple active learning scheme, which can be summarised as\footnote{For a detailed description of the algorithm used, we refer to Section~\ref{ala} and Algorithm~\ref{algo:active_loop}.}
\begin{enumerate}
    \item Train a neural network classifier on your imbalanced training set, using normal cross-entropy loss.
    \item From the unlabeled set, select the frames with the highest probability of belonging to one of the rare classes, as estimated by the classifier. Add these to the training set.
\end{enumerate}
We were surprised that this simple algorithm worked so well, since the large class imbalance leads to a very low estimated probability for most of the rare classes, even for the selected samples. To explain this surprising result we analyse a simple 2-dimensional toy model of a dataset with high class imbalance.

It is possible that one could improve our results further by making use of different techniques that have been employed to improve the performance of neural network classifiers under strong class imbalance. In the Appendix we try to incorporate two commonly used few-shot learning algorithms into our method, but we find that the added complexity does not lead to improved results. However, there are many more few-shot learning methods that we did not try \citep{hu_leveraging_2021}, as well as other techniques such as loss reweighting.

Finally, in this paper we  limited ourselves to the problem of traffic sign classification. We believe that a similar method could work well for other classification problems as well, yet it is worth keeping in mind that traffic signs are special. In contrast with naturally occuring objects such as different animal and plant species, traffic signs are designed to be easily distinguishable by a neural network (the human visual cortex). This could be the reason why even rare classes can be rather well separated from the other classes, even when training on just one or a few samples from the rare class.

\hfill \break

{\em Acknowledgments}. The results presented in this manuscript were obtained by S. Jaghouar and H. Gustafsson in their joint Master project  \citep{Jaghouar,Gustafsson}.
This work would not have been possible without the help of Mahshid Majd and her expertise on traffic sign recognition.

\bibliographystyle{unsrt}

\appendix
\section{Feature space dissection}
\label{sec:feature_space}

\begin{figure}[t]
	\includegraphics[width=0.4\textwidth,]{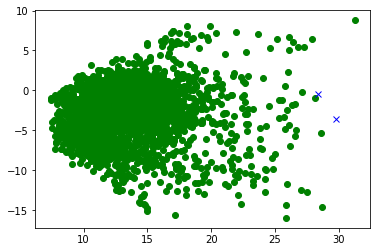}
	\caption{A two-dimensional projection of the 512-dimensional feature space of the ResNet model trained on traffic signs. To obtain a projection that brings out the difference between the rare class and the others, we  performed uncentered PCA for the 6 {\it training} samples of the rare class, and used this projection to map out the feature space of the {\it validation} set. The two blue crosses are the samples of the same rare class, and the green circles are all the other samples in the validation set. Note that despite the large class imbalance, the two samples corresponding to the rare class are among the four rightmost points.
	}
	\label{fig:pca}

\end{figure}

{\em Feature-space dissection.} We  discussed in Section~\ref{discussion} how it can be possible for a standard neural network, trained with unweighted cross-entropy loss, to identify samples from the rare class with such high precision, despite a low estimated probability.
However, this requires first that the feature space is expressive enough that the training samples of the rare class share a consistent direction in feature space, second that unseen samples from this class are also preferentially found in this direction.
Fig.~\ref{fig:pca} shows that this is indeed the case for our trained ResNet model.
In this figure, we show a two-dimensional linear projection of the feature space, which has been designed to bring out the dimension along which the training samples belonging to the rare class differ the most from the other samples.
This is the $x$-axis of Fig.~\ref{fig:pca}.
In this figure, we see the projection of the {\it validation} set.
The two samples from the rare class that are present in this set are very far to the right.
This indicates that the direction defined by the average of the training samples of the rare class is a good direction to search in, if one wants to find unseen samples from this class.
Following the discussion above, this also indicates that we can expect good results by using the output probability of this class to search for samples.
This is exactly what we found in Section~\ref{results}.

\section{ Comparison with few-shot learning}
\label{app:few_shot}

{\em Comparison with few-shot learning}. The efficiency of our algorithm is explained by the fact that it recognizes rare traffic signs after  seeing only a single rare sample amongst a large number of common traffic signs. 
The algorithm works so well because it  uses features learned from the common traffic signs  to identify rare ones.
This is quite similar to the problem of few-shot learning. We therefore tested whether we could improve the performance of our active-search algorithm by using few-shot learning methods when training the network used for searching. We tested two few-shot learning methods, ProtoNet \citep{snell_prototypical_2017} and RelationNet \citep{sung_learning_2018}. For each active learning iteration, we trained a network using the respective method, and used it for selecting what frames to move to the training set. RelationNet and ProtoNet were trained in a one-shot scenario with training episodes containing 50 classes and 8 queries. ProtoNet and RelationNet take a query sample and one or more support samples as input. ProtoNet predicts the distance between the query and the support samples, and RelationNet the similarity between the two. Therefore, to perform the search, the training samples of a given rare class are used as support samples, while each sample in the unlabeled set is used as a query to compute a distance/similarity between the two. For ProtoNet, the samples with the smallest distance are moved to the training set. For RelationNet, the samples with highest similarity are moved.
To evaluate the performance of these methods, a normal ResNet model was trained on the new training set, and used to calculate the $F_1$ score.
To get a fair comparison, we used the same ResNet backbone pretrained on ImageNet for both few-shot models. Our implementations are available in a \href{https://github.com/samsja/finding-a-needle-code}{github repository}.

\begin{figure}[b]
	\begin{overpic}[scale=0.6,angle =0]{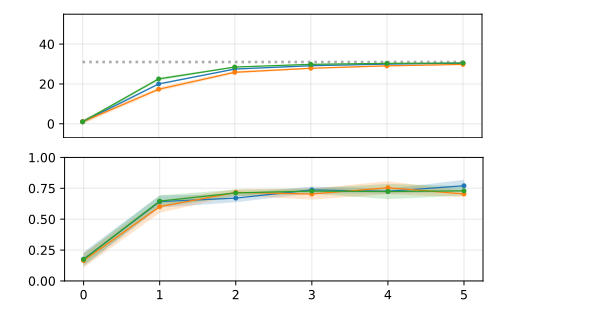}
	\put(-2,40){\colorbox{white}{ \footnotesize $N_{\rm rare}$}}
	\put(-2,17){\colorbox{white}{ \footnotesize $F_1$}}
	\put(62,-1){\colorbox{white}{ \footnotesize $t$}}
        \put(12,44){\colorbox{white}{({\bf a})}}
        \put(12,20){\colorbox{white}{({\bf b})}}
	\end{overpic}
	\caption{Performance of the active-search algorithm. ({\bf a}) The number $N_{\rm rare}$ of training samples per class, averaged over the rare classes in Fig.~\ref{fig:25_rarest_class}, as a function of the number of iterations $t$ in the active loop. The dotted gray line in panel ({\bf a}) shows the maximal value, i.e. when all the rare samples have been moved to the training set. Training methods used: Standard cross-entropy training (green), ProtoNet (blue), RelationNet (orange). See the text for further details. Each experiment was conducted five times, and the corresponding 95\% confidence intervals are shown as shaded areas.  ({\bf b}) Same, but for $F_1$~score.}
	\label{fig:al_few_shot}
	\centering
\end{figure}
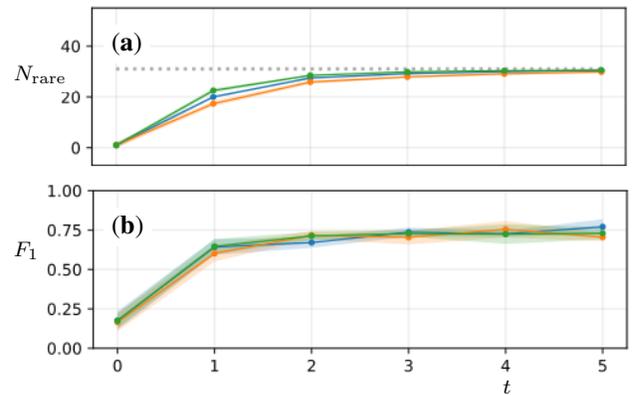

In Fig.~\ref{fig:al_few_shot}, we compare the performance of our active-search algorithm with and without the use of few-shot methods. In Fig.~\ref{fig:al_few_shot}({\bf a}), we show that the few-shot methods do a little bit worse than the standard training method, when it comes to how many samples from the rare classes that they manage to identify. In Fig.~\ref{fig:al_few_shot}({\bf b}), we show the performance ($F_1$~score) of a classification model trained on the resulting training sets. Here, we see no clear difference between the models, when it comes to how well they can classify rare classes. In sum, there is no benefit from incorporating these techniques into our active-search algorithm.

\end{document}